\documentclass[letterpaper, 10 pt, conference]{ieeeconf}  
\IEEEoverridecommandlockouts                              
\usepackage{graphics} 
\usepackage{epsfig} 
\usepackage{mathptmx} 
\usepackage{times} 
\usepackage{amsmath} 
\usepackage{amssymb}  
\usepackage{hyperref}
\usepackage{url}
\usepackage{multirow}
\usepackage{gensymb}
\usepackage[utf8x]{inputenc}
\hypersetup{
    colorlinks=true,
    linkcolor=blue,
    filecolor=magenta,      
    urlcolor=cyan,
}

\title{\LARGE \bf
Are We Ready for Service Robots? The OpenLORIS-Scene Datasets for Lifelong SLAM
}

\author{Xuesong Shi$^{*1}$, Dongjiang Li$^{*2,4}$, Pengpeng Zhao$^{1,5}$, Qinbin Tian$^{2,4}$, Yuxin Tian$^{2,5}$, Qiwei Long$^{2,4}$,\\
Chunhao Zhu$^{2,4}$, Jingwei Song$^{2,4}$, Fei Qiao$^{2}$, Le Song$^{3}$, Yangquan Guo$^{3}$, Zhigang Wang$^{1}$,\\
Yimin Zhang$^{1}$, Baoxing Qin$^{3}$, Wei Yang$^{4}$, Fangshi Wang$^{4}$, Rosa H. M. Chan$^{6}$ and Qi She$^{1}$
\thanks{*Equal contribution.}
\thanks{$^{1}$Intel Labs China, Beijing, 100190 China.}%
\thanks{$^{2}$Department of Electronic Engineering and BNRist, Tsinghua University, Beijing, 100084 China.}%
\thanks{$^{3}$Gaussian Robotics, Shanghai, 201203 China.}%
\thanks{$^{4}$Beijing Jiaotong University, Beijing, 100044 China.}%
\thanks{$^{5}$Beihang University, Beijing, 100191 China.}%
\thanks{$^{6}$City University of Hong Kong, Hong Kong, China.}%
\thanks{Corresponding authors: 
    {\tt\small xuesong.shi@intel.com},
    {\tt\small qiaofei@tsinghua.edu.cn}.}
}

\begin{document}

\maketitle

\begin{abstract}

Service robots should be able to operate autonomously in dynamic and daily changing environments over an extended period of time. While Simultaneous Localization And Mapping (SLAM) is one of the most fundamental problems for robotic autonomy, most existing SLAM works are evaluated with data sequences that are recorded in a short period of time. In real-world deployment, there can be out-of-sight scene changes caused by both natural factors and human activities. For example, in home scenarios, most objects may be movable, replaceable or deformable, and the visual features of the same place may be significantly different in some successive days. Such out-of-sight dynamics pose great challenges to the robustness of pose estimation, and hence a robot's long-term deployment and operation. To differentiate the forementioned problem from the conventional works which are usually evaluated in a static setting in a single run, the term \textit{lifelong SLAM} is used here to address SLAM problems in an ever-changing environment over a long period of time. To accelerate lifelong SLAM research, we release the OpenLORIS-Scene datasets. The data are collected in real-world indoor scenes, for multiple times in each place to include scene changes in real life. We also design benchmarking metrics for lifelong SLAM, with which the robustness and accuracy of pose estimation are evaluated separately. The datasets and benchmark are available online at \href{https://lifelong-robotic-vision.github.io/dataset/scene}{lifelong-robotic-vision.github.io/dataset/scene}.

\end{abstract}

\begin{figure*}[t!]
\centerline{\includegraphics[width=18cm]{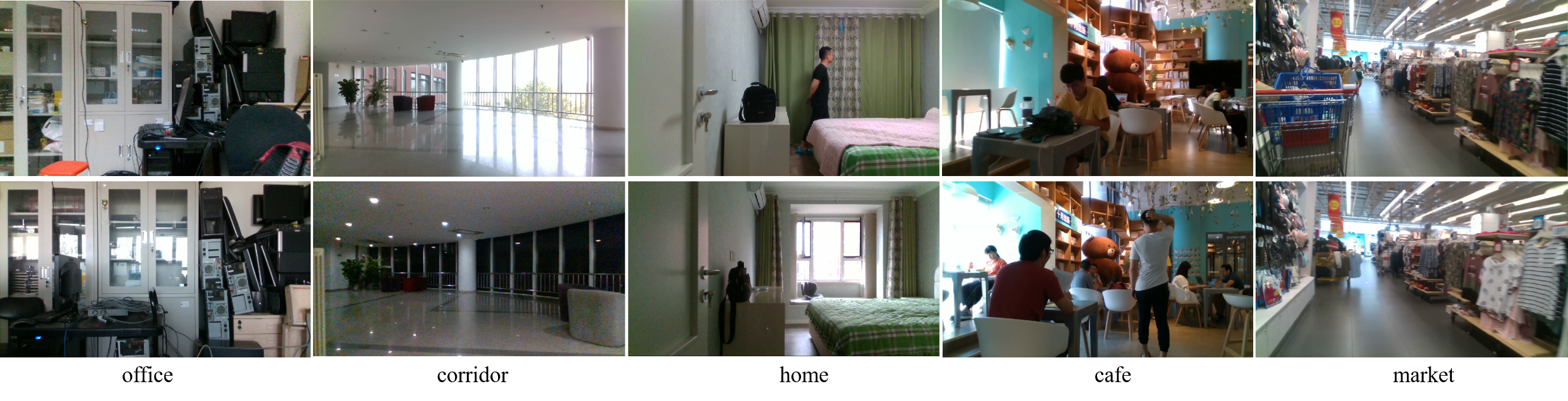}}
\caption{Examples of color images in the OpenLORIS-Scene datasets. The upper and lower images in each column show approximately the same place in different data sequences, but the scene had been changed.}
\label{fig_scenes}
\end{figure*}

\section{Introduction}

The capability of continuous self localization is fundamental to autonomous service robots. Visual Simultaneous Localization and Mapping (SLAM) has been proposed and studied for decades in robotics and computer vision. There have been a number of open source SLAM systems with careful designs and heavily optimized implementations. Do they suffice for deployment in real-world robots? We claim there is still a gap, coming from the fact that most SLAM systems are designed and evaluated for a single operation. That is, a robot moves through a region, large or small, with a fresh start. Real-world service robots, on the contrary, usually need to operate at a region day after day, with the requirement of reusing a persistent map in each operation to retain spatial knowledge and coordinate consistency. This requirement is more than saving the map and loading it for the next operation. The scene changes in real life and other uncontrolled factors in a long-term deployment bring considerable challenges to SLAM algorithms.

In this work, we use the term \textit{lifelong SLAM} to describe the SLAM problem in long-term robot deployments. \textit{For a robot that needs to operate around a particular region over an extended period of time, the capability of lifelong SLAM aims to build and maintain a persistent map of this region and to continuously locate the robot itself in the map during its operations.} To this end, the map must be reused in different operations, even if there are changes in the environment.

We summarize the major source of algorithmic challenges for lifelong SLAM as following:

\begin{itemize}
\item \textit{Changed viewpoints} - the robot may see the same objects or scene from different directions.
\item \textit{Changed things} - objects and other things may have been changed when the robot re-enters a previously observed area.
\item \textit{Changed illumination} - the illumination may change dramatically. 
\item \textit{Dynamic objects} - there may be moving or deforming objects in the scene.
\item \textit{Degraded sensors} - there may be unpredictable sensor noises and out-of-calibrations due to mechanical stress, temperature change, dirty or wet lens, \textit{etc}.
\end{itemize}

While each of these challenges has been more or less addressed in existing works, there is a lack of public datasets and benchmarks to unify the efforts towards building practical lifelong SLAM systems. Therefore, we introduce the OpenLORIS-Scene datasets, which are particularly built for the research of lifelong SLAM for service robots. The data are collected with commodity sensors carried by a wheeled robot in typical indoor environments as shown in Fig. \ref{fig_scenes}. Ground-truth robot poses are provided based on either a Motion Capture System (MCS) or a high-accuracy LiDAR. The major distinctions of our datasets are:

\begin{itemize}
\item The data are from real-world scenes with people in it.
\item There are multiple data sequences for each scene, which include not only changes in illumination and viewpoints, but also scene changes caused by human activities in their real life.
\item There is a rich combination of sensors including RGB-D, stereo fisheyes, inertial measurement units (IMUs), wheel odometry and LiDAR, which can facilitate comparison between algorithms with different types of inputs.
\end{itemize}

This work also proposes new metrics to evaluate lifelong SLAM algorithms. As we believe the robustness of localization should be the most important concern, we use correct rates to explicitly evaluate it, as opposed to existing benchmarks where robustness is partially implied by the accuracy metrics.


\section{Related Works}

The adjective of \textit{lifelong} has been used in SLAM-related works to emphasis either or both of the two capabilities: robustness against scene changes, and scalability in the long run. A survey of both directions can be found in \cite{cadena2016past}.

Most SLAM works evaluate their algorithms on one or more public datasets to justify their effectiveness in certain aspects. The most well-used datasets include TUM RGB-D \cite{tum-rgbd}, EuRoC MAV \cite{euroc} and KITTI \cite{kitti}. A recent contribution is the TUM VI benchmark \cite{tum-vi}, where aligned visual and IMU data are provided. One of the major distinctions of those datasets is their sensor types. While there is favor of RGB-D data source in recent SLAM algorithm research for dense scene reconstruction, there is a lack of dataset with both RGB-D and IMU data. Our dataset provides aligned RGB-D-IMU data, along with odometry data which are widely used in the industry but often lack in public datasets.

Synthesized datasets are also used for SLAM evaluation. Recent progress in random scene generation and photo-realistic rendering \cite{mccormac2016scenenet}\cite{li2018interiornet} makes it theoretically possible to synthesize scene changes for lifelong SLAM, but it would be difficult to model realistic changes as in natural lives.

For real-world scene changes, the COLD database \cite{cold} provides visual data of several scenes with variations caused by weather, illumination, and human activities. It is the most related work with ours in the principle of data collection, though with different sensor setups. Object-level variations can also be found in the change detection datasets \cite{fehr2017change-detection}, but it is not designed for SLAM and does not provide ground-truth camera poses.

Recently there are efforts towards unified SLAM benchmarking and automatic parameter tuning \cite{Bodin2018slambench2}\cite{gslam2019}, our work contributes to this direction by introducing new data and performance metrics.

\section{OpenLORIS-Scene Datasets}

The OpenLORIS-Scene datasets are designed to be a testbed of real-world practicality of lifelong SLAM algorithms for service robots. Therefore, the major principle is to make the data as close to service robots scenarios as possible. Commercial wheeled robot models equipped with commodity sensors are used to collect data in typical indoor scenes with people in it, as shown in Fig. \ref{fig_scenes}. Rich types of data are provided to enable comparison of methods with different kind of inputs, as listed in Table \ref{tab_sensors}. All the data are calibrated and synchronized.

\subsection{Robots and Sensors}

To enable monocular, stereo, RGB-D and visual-inertial SLAM algorithms, two camera devices are used for data collection: a RealSense D435i providing RGB-D images and IMU measurements, and a RealSense T265 tracking module providing stereo fisheye images and IMU measurements. The IMU data are hardware synchronized with images from the same device. The cameras are mounted on a customized Segway Deliverybot S1 robot (for all the scenes except \texttt{market}) or a Gaussian Scrubber 75 robot (for the \texttt{market} scene), front-facing, at the height of about one meter. The resolution of RGB-D images are chosen to maximize the field of view (FOV), and for the best depth quality \cite{realsense_bkm}. We provide not only aligned depth data as in other RGB-D datasets, but also raw depth images since they have a larger FOV which could benefit depth-based SLAM algorithms.

Wheel encoder-based odometry data are also provided, as they are widely available in wheeled robots. The odometry data from the Segway robot are fused from wheel encoders and a chassis IMU by proprietary filtering algorithms along with the robot.

To provide ground-truth trajectories, the Segway robot equips markers of an OptiTrack MCS and a Hokuyo UTM-30LX LiDAR, all near the cameras. The Gaussian robot equips a RoboSense RS-LiDAR-16.

\begin{table}[htbp]
\caption{Data Types in the OpenLORIS-Scene Datasets}
\centering
\begin{tabular}{ccccc}
\hline
\textbf{Device}& \textbf{Data}& \textbf{FPS}& \textbf{Resolution} & \textbf{FOV} \\
\hline
D435i &	color	& 30 & 848x480 & H:69 V:42 D:77\\
D435i & depth	& 30 & 848x480 & H:91 V:65 D:100\\
D435i & aligned depth$^{\mathrm{a}}$	& 30 & 848x480 & H:69 V:42 D:77\\
D435i & accel	& 250 & - & -\\
D435i & gyro	& 400 & - & -\\
T265  & fisheye1	& 30 & 848x800   & D:163\\
T265  & fisheye2	& 30 & 848x800   & D:163\\
T265  & accel   & 62.5 & - & -\\
T265  & gyro    & 200 & - & -\\
base  & odometry	& 20$^{\mathrm{b}}$ & - & -\\
LiDAR & laser scan & 40 & 1080 & H:270$^{\mathrm{c}}$\\
\hline
\multicolumn{5}{l}{$^{\mathrm{a}}$ Depth images aligned to color images for per-pixel correspondence.}\\
\multicolumn{5}{l}{$^{\mathrm{b}}$ This value is different for the \texttt{market} scene: 40.} \\
\multicolumn{5}{l}{$^{\mathrm{c}}$ This value is different for the \texttt{market} scene: H:360 V:30.}
\end{tabular}
\label{tab_sensors}
\end{table}

\begin{table}[htbp!]
\caption{Tools Used for Extrinsic Calibration}
\centering
\begin{tabular}{ccc}
\hline
\multicolumn{2}{c}{\textbf{Sensors}} & \textbf{Tool} \\
\hline
D435i & T265 & Kalibr \cite{kalibr} \href{https://github.com/ethz-asl/kalibr}{github.com/ethz-asl/kalibr} \\
\hline
\multirow{2}{2em}{MCS} & \multirow{2}{2.4em}{D435i} & robot\_cal\_tools \\
& & \href{https://github.com/Jmeyer1292/robot\_cal\_tools}{github.com/Jmeyer1292/robot\_cal\_tools} \\
\hline
MCS & T265 & basalt \cite{tum-vi} \href{https://gitlab.com/VladyslavUsenko/basalt}{gitlab.com/VladyslavUsenko/basalt} \\
\hline
\multirow{2}{3em}{LiDAR} & D435i & LaserCamCal \cite{lasercamcal} \\
& T265 & \href{https://github.com/MegviiRobot/CamLaserCalibraTool}{github.com/MegviiRobot/CamLaserCalibraTool} \\
\hline
\multirow{2}{4em}{odometer} & D435i & \multirow{2}{5em}{proprietary} \\
& T265 &  \\
\hline
\multicolumn{3}{l}{D435i refers to its color camera; T265 refers to its left fisheye camera.}
\end{tabular}
\label{tab_calibration}
\end{table}

\vspace*{-10pt}

\subsection{Calibration}

The intrinsics and intra-device extrinsics of cameras and IMUs are from factory calibration. Other extrinsics are calibrated with the tools listed in Table \ref{tab_calibration}. Redundant calibrations are made for quality evaluation. Each non-camera sensor (MCS, LiDAR and odometer) is calibrated against both cameras, so that the extrinsics between the two cameras can be deduced, which is then compared with their extrinsics directly calibrated with Kalibr \cite{kalibr}. The resulted errors are all below 1cm in translation and {2\degree} in rotation, except for odometry calibration whose translation error is 7cm.

\subsection{Synchronization}

Images and IMU measurements from the same RealSense device are hardware synchronized. Software synchronization is performed for each data sequence between data from different devices, including RealSense D435i, RealSense T265, LiDAR, MCS and odometer. For each of those devices, its trajectory can be obtained either via a SLAM algorithm or directly from the measurements. Those per-device trajectories are then synchronized by finding the optimal time offsets to minimize the RMSE of absolute trajectory errors (ATEs). The ATEs of each per-device trajectory are calculated against the trajectory of MCS for the scene of {\verb office }, and T265 for others, as the two provide poses in highest rates.

To mitigate the affection by SLAM and measurement noises, we generated a controlled piece of data at the beginning of each data sequence by pushing the robot back and forth for several times in a static and feature rich area, and used only this piece of data for synchronization.

The synchronization quality is evaluated by the consistency of resulted optimal time offsets. From our experiments, the standard deviation of offsets ranges from 1.7 ms (MCS to T265) to 7.4 ms (odometry to T265), with a positive correlation with the measurement cycle of each sensor. We think the results are acceptable for our scenarios, yet better synchronization methods can be discussed. One inherent drawback of the ATE minimization method is that systematic errors can be introduced if the scale of each estimated trajectory differs, which is frequently observed in the data. We mitigate this effect by using back-and-forth trajectories instead of move-and-stop ones, and also by carefully selecting a period of data when all trajectories can be best matched.

\subsection{Scenes and Sequences}

There are five scenes in the current datasets. For each scene, there are 2-7 data sequences recorded at different times. The sequences are manually selected and clipped from much more recordings to form a concise benchmark including most major challenges in lifelong SLAM. Some of the scene changes were deliberately influenced by the authors to maximize the difference between sequences, but all the manual changes were those that would likely to happen given a longer time. (e.g. relocated table and sofa in corridor)

\begin{itemize}
\item {\verb Office }: 7 sequences in a university office with benches and cubicles. This scene is controlled: in \texttt{office-1} the robot walked along a U-shape route; in {\verb office-2 } the scene is unchanged but the route is reversed, so the cameras observe from opposite views; {\verb office-3 } is a turn-around that could connect {\verb office-1 } and {\verb office-2 } (to potentially be used to align sub-maps); in {\verb office-4 } and {\verb office-5 } the illumination is different from previous sequences; in {\verb office-6 } there are object changes; and {\verb office-7 } further introduced dynamic objects (persons).

\item {\verb Corridor }: 5 sequences in a long corridor with a lobby in the middle and the above office at one end. Apart from the well-known challenges in feature-poor long corridors, additional difficulties come from the high contrast between the corridor and the window at daytime, and extremely low light at night. Between sequences, there are not only illumination changes, but also moved furniture, which could make re-localization and loop closure a tough task. And the largeness of the scene would magnify the inconsistency of maps from different sequences if the algorithm fails to align them.

\item {\verb Home }: 5 sequences in a two bedroom apartment. There are lots of scene differences between sequences, such as changed sheets and curtains, moved sofa and chairs, and people moving around.

\item {\verb Cafe }: 2 sequences in an open café. There are different people and different things in each sequence. 

\item {\verb Market }: 3 sequences in an open supermarket. Each trajectory is a long loop (150-220 meters). There are people moving around in the scene. The goods on some shelves have been changed between sequences.
\end{itemize}

There are 22 sequences in total. The accumulated length of the data is 2244 seconds.

\subsection{Ground-truth}

For each scene, ground-truth robot poses in a persistent map are provided for all sequences. For the {\verb office } scene they are obtained from an MCS which wholly covers all the sequences, with a persistent coordinate system. The MCS-based ground-truth is in a rate of 240 Hz, with outliers removed. For other scenes, a 2D laser SLAM method is employed to generate ground-truth poses. A full map is built for each scene, and the robot is localized in the map with each frame of laser scan in the sequences. For the scene of {\verb corridor } and {\verb cafe }, a variant of hector\_mapping \cite{hector} is used for map construction and localization. For {\verb home } and {\verb market }, another laser-based SLAM system combined with multi-sensor fusion is used to avoid from mismatching. The initial pose estimation of each sequence is manually assigned, and the output is manually verified to be correct. A comparison between laser-based ground-truth and MCS-based ground-truth is made with the in-office part of {\verb corridor } data, which gives an ATE of 3 cm.

\section{Benchmark Metrics}

Like most existing SLAM benchmarks, we mainly evaluate the quality of camera trajectory estimated by the SLAM algorithms. We adopt the same definition of Absolute trajectory error (ATE) and Relative pose error (RPE) as in the TUM RGB-D benchmark \cite{tum-rgbd} to evaluate the accuracy of pose estimation for each frame. However, estimation failures or wrong (mismatched) poses are more severe than inaccuracies, and they may occur more commonly in lifelong SLAM due to scene changes. Therefore, we design separate metrics to evaluate the correctness and accuracy respectively.

\subsection{Robustness Metrics}

\textit{Correctness}. For each pose estimate $p_k$ at time $t_k$, given the ground-truth pose at that time, we assess the correctness of the estimate by its ATE and absolute orientation error (AOE):

\begin{equation}
c^{\epsilon,\phi}(p_k)=
\begin{cases}
1, & \text{if}\ \text{ATE}\left(p_{k}\right)\leq\epsilon \text{ and } \text{AOE}(p_k)\leq\phi \\
0, & \text{otherwise}
\end{cases}
\end{equation}

\textit{Correct Rate (CR) and Correct Rate of Tracking (CR-T)}. While correctness evaluates a single pose estimate, the overall robustness metric over one or more data sequences can be defined as the correct rate over the whole time span of data. For a sequence from $t_{\min }$ to $t_{\max }$, given an estimated trajectory $\{t_k, p_k\}_{k=0,...,N}$, define
\begin{equation}
\text{CR}^{\epsilon,\phi}=\frac{\sum_{k=0}^{N} \left(\min \left(t_{k+1}-t_{k}, \delta\right) \cdot c^{\epsilon,\phi}(p_{k})\right)}{t_{\max }-t_{\min}},
\end{equation}
\begin{equation}
\text{CR}^{\epsilon,\phi}\text{-T}=\frac{\sum_{k=0}^{N} \left(\min \left(t_{k+1}-t_{k}, \delta\right) \cdot c^{\epsilon,\phi}(p_{k})\right)}{t_{\max }-t_{0}},
\end{equation}
where $t_{N+1}\doteq t_{\max}$, $\delta$ is a parameter to determine how long a correct pose estimation is valid for. Note that in $\text{CR}^{\epsilon,\phi}\text{-T}$ the time for re-localization and algorithm initialization ($t_0-t_{\min}$) is excluded, since tracking is not functioning during that time. In practice, the ATE threshold $\epsilon$ and AOE threshold $\phi$ should be set according to the area of the scene and the expected drift of the SLAM algorithm. $\delta$ should be set larger than the normal cycle of pose estimation, and much smaller than the time span of data sequence. For common room or building size data, we would suggest to set $\epsilon$ to meter-size and $\delta$ around one second.

\textit{Correctness Score of Re-localization (CS-R)}. As tracking and re-localization are often implemented with different methods in common SLAM pipelines, they should be evaluated separately. The correctness of re-localization can be decided by the same ATE threshold as in CR. But besides correctness, we would also like to know how much time it takes to re-localize. Therefore, we define a score of re-localization as
\begin{equation}
\text{C}^{\epsilon,\phi} \text{S}^{\tau}\text{-R}=
e^{-{(t_0-t_{\min})/\tau}}\cdot c^{\epsilon,\phi}(p_0)
\end{equation}
where $\tau$ is a scaling factor. Note that for an immediate correct re-localization with $t_0=t_{\min}$, there will be $\text{CS-R}=1$. The score drops with the time for re-localization increases. For normal evaluation cases we would suggest to set $\tau=60 \text{s}$.

\subsection{Accuracy Metrics}

To evaluate the accuracy of pose estimation without affected by incorrect results, we suggest to use statistics of ATE and RPE over one or more trajectories with only correct estimations. For example, C$^{0.1}$-RPE RMSE is the root mean square error of RPE of correct pose estimates selected by an ATE threshold of 0.1 meter.


\section{Experiments}

\begin{figure*}[t!]
\centerline{\includegraphics[width=17.5cm,trim={0.1cm 1.2cm 3cm 1cm},clip]{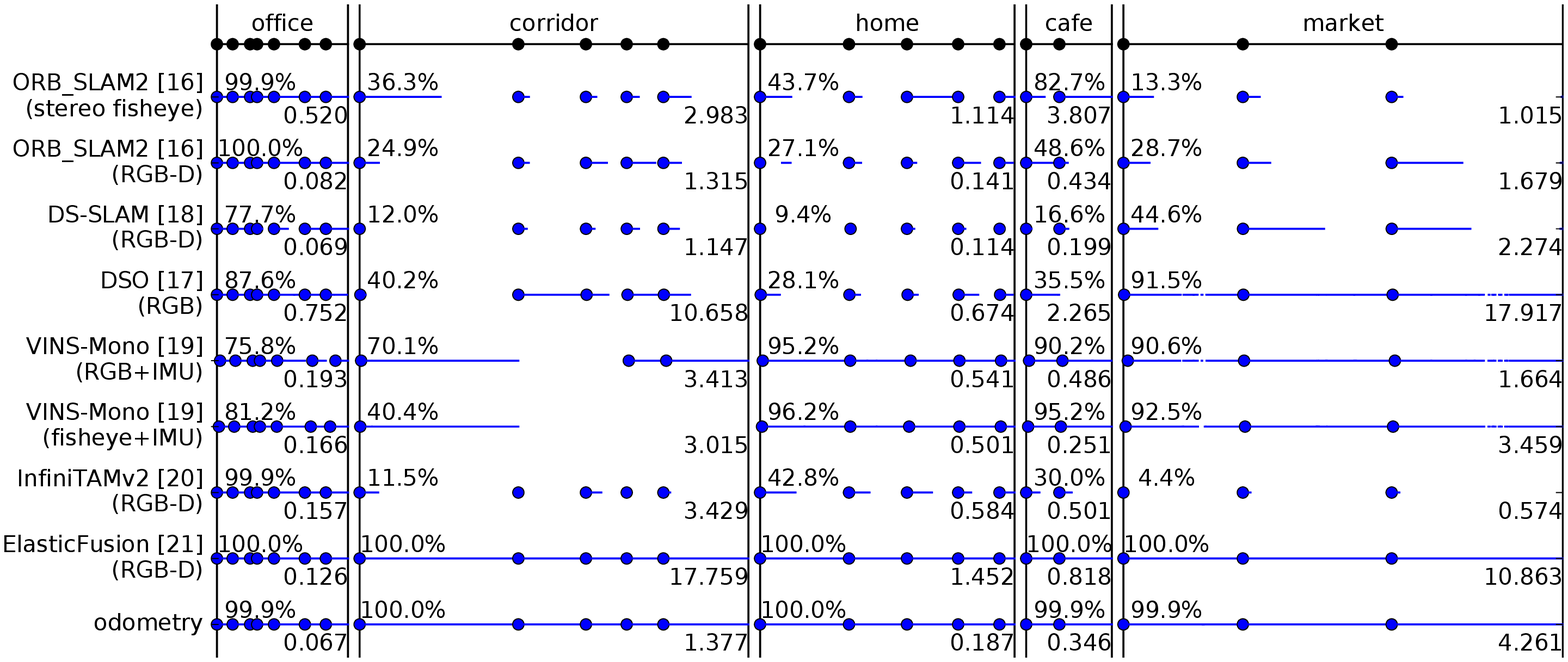}}
\caption{Per-sequence testing results with the OpenLORIS-Scene datasets. Each black dot on the top line represents the start of one data sequence. For each algorithm, blue dots indicate successful initialization, and blue lines indicate successful tracking. The percentage value on the top left of each scene is average CR$^{\infty}$, larger means more robust. The float value on the bottom right is average ATE RMSE, smaller means more accurate.} \label{fig_per_seq}
\end{figure*}

\begin{figure*}[t!]
\centerline{\includegraphics[width=17.5cm,trim={0.1cm 1.2cm 3cm 1cm},clip]{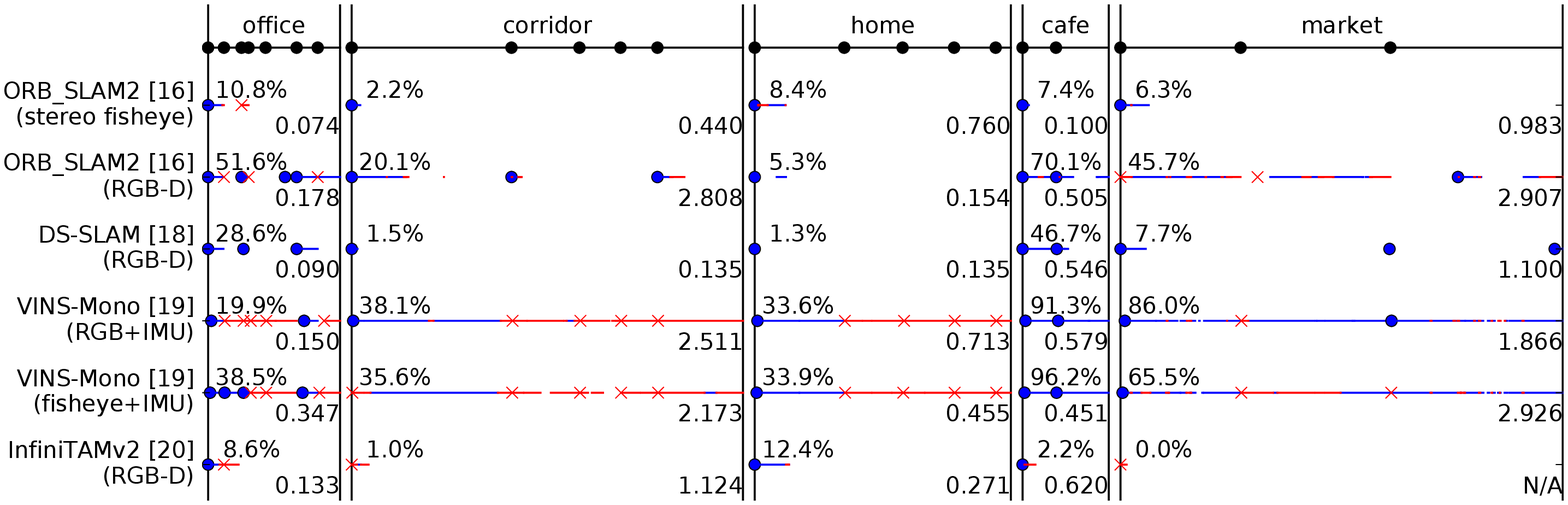}}
\caption{Lifelong SLAM testing results with the OpenLORIS-Scene datasets. For each algorithm, blue dots indicate successful initialization or correct re-localization, while red crosses are incorrect re-localization. Line segments in blue and red indicate correct and incorrect pose estimation, respectively. The percentage value on the top left of each scene is average Correct Rate (CR$^{\epsilon,\phi}$ as in Eq. (2)). The float value on the bottom right is average C$^{\epsilon,\phi}$-ATE RMSE. The ATE threshold $\epsilon$ is 1 m for office, 3 m for home and cafe, and 5 m for corridor and market. The AOE threshold $\phi$ is {30\degree} for all scenes.} \label{fig_lifelong}
\end{figure*}

The OpenLORIS-Scene datasets and the proposed metrics are tested with open-source SLAM algorithms. The algorithms are chosen to cover most data types listed in Table \ref{tab_sensors}, and to represent a diverse set of SLAM techniques. {ORB-SLAM2} is a feature-based SLAM algorithm \cite{orbslam2}. It can optimize poses with absolute scale by using either stereo features or depth measurements. DSO, on the contrary, tracks the camera's states with a fully direct probabilistic model \cite{dso}. DS-SLAM improves over ORB-SLAM2 by removing features on moving objects \cite{ds-slam}. VINS-Mono provides robust pose estimates with absolute scale by fusing pre-integrated IMU measurements and feature observations \cite{vins-mono}. InfiniTAM is a dense SLAM system based on point cloud matching with an iterative closed point (ICP) algorithm \cite{infinitamv2}. ElasticFusion combines the merits of dense reconstruction and globally consistent mapping by using a deformable model \cite{efusion}.

\subsection{Per-sequence Evaluation}

\textit{Method.} First we test each data sequence separately, as done in most existing works. For each algorithm, the ground-truth trajectory are transformed into the target frame of pose estimation, for example, the color sensor of D435i for ORB-SLAM2 with RGB-D input. Then the estimated trajectory are aligned with the ground-truth using the method of Horn. For DSO, an optimal scaling factor is calculated with Umeyama's method \cite{umeyama}. Then ATE of each matched pose is calculated and their RMSE over each sequence is reported. The only difference in our ATE calculation process from conventional ones is that we interpolate the ground-truth trajectory to let each estimated pose get an exact match on the timeline, as opposed to matching the closest ground-truth pose. The reason is that our laser-based ground-truth trajectories have a lower rate than MCS-based ones.

\textit{Result.} The results are visualized in Fig. \ref{fig_per_seq}, with blue line segments indicating successful localization and blank otherwise. The success rate indicated by CR$^{\infty}$, and accuracy indicated by ATE RMSE are calculated for each sequence. On the figure only statistics over each scene are shown, where CR$^{\infty}$ and ATE RMSE are averaged weighted by the time span of each sequence and the count of pose estimates, respectively. All the algorithms can track successfully most of the time in {\verb office }, but other scenes are challenging. For example, most algorithms tend to lost in {\verb corridor } because of the featureless walls and low light, yet VINS-Mono can fully track some of the sequences in this scene. Note that VINS-Mono fails to initialize in some low-light sequences in {\verb corridor }. Nevertheless, VINS-Mono shows the best robustness among the tested algorithms.

The wheel odometry data in OpenLORIS-Scene are evaluated along with SLAM algorithms in Fig. \ref{fig_per_seq}. It can be seen that our odometry data provides reliable tracking results even in large scenes. We think that odometry should not be neglected by practical SLAM algorithm designers for service robots.

\textit{Metrics discussion.} If we compare between the CR$^{\infty}$ from DS-SLAM and ORB-SLAM2 with the same inputs, the former tends to lost more often since it uses less features to localize, but it succeeds in {\verb market } which are highly dynamic. If we also note their ATEs, it can be found a consistent negative correlation between the two similar algorithms' ATE and CR$^{\infty}$. The reason is that the longer an algorithm tracked, the more error is likely to accumulate. It implies that evaluating algorithms purely by ATE could be misleading. On the other hand, considering only CR$^{\infty}$ could also be misleading. For example, DSO results high CR$^{\infty}$ in {\verb corridor } and {\verb market }, but the estimated trajectories are actually erroneous. Its CR would be much lower if we set a proper ATE threshold.

\subsection{Lifelong SLAM Evaluation}

\textit{Method.} To test whether an algorithms could continuously localize in changed scenes, we feed it sequences of the same scene one by one. There may be a significant view change when switching to the next sequence. The algorithm could either wait for a successful re-localization (e.g. ORB-SLAM2), or start with a fresh map and then try to align it with the old map by loop closing (e.g. VINS-Mono). DSO and ElasticFusion are excluded from this test since the implementation we use does not support re-localization. For ORB-SLAM2 RGB-D, we use a revised version with a few engineering improvements but no algorithmic changes. For each scene, we align the estimated trajectory of the first sequence to the ground-truth, and using the resulted transformation matrix to transform all the estimated trajectories of this scene, then compare them with the ground-truth.

\textit{Result.} The results are shown in Fig. \ref{fig_lifelong}, with red cross and line segment indicating incorrect pose estimates, judged by an ATE threshold of 1/3/5 meters for small/medium/large scenes and an AOE threshold of {30\degree}. It shows that re-localization is challenging. For example, most algorithms completely fail to re-localize in the 2\textsuperscript{nd}-5\textsuperscript{th} sequences of {\verb home }.

\textit{Metrics discussion.} From the results we see that the metrics are imperfect. For example, for {\verb corridor } and {\verb market }, some algorithms get an incorrect initial localization for the first sequence, which is technically unsound. The reason is that large drifts have been accumulated over the long trajectories, and after aligning the full trajectory to the ground-truth, its initial part has a large error. It suggests that we should set even larger ATE thresholds for large scenes, and that further refinement of the accuracy judgement method should be discussed. Besides the false alarm in initial and final parts of {\verb corridor-1 } and {\verb market-1 }, the metrics succeeds to recognize incorrect localization, and gives meaningful statistics.

\textit{Factor analysis}. Correct re-localization is rare in Fig. \ref{fig_lifelong} partly because we have deliberately selected the most challenging sequences for the released data. In most scenes, the challenge comes from mixed factors including changed viewpoints, changed illumination, changed things and dynamic objects. The {\verb office } data are designed to disentangle those factors. We conduct another set of tests with specified sequence pairs in {\verb office }. The two sequences in each pair have one key different factors, as described in Section III.D. The re-localization scores are listed in Table \ref{tab_reloc}. The results suggest that changed viewpoints and illumination are most difficult to deal with. The former is expected as natural scenes are likely to generate different visual and geometric features from different viewpoints. The latter might be mitigated by carefully tuning algorithms and devices. We expect that deep learning based features and semantic information should be able to help address both problems.

\begin{table}[tb!]
\centering
\caption{Re-localization Scores with Controlled Changing Factors}
\begin{tabular}{ccccccc}
\hline
\textbf{Data: office-}& \textbf{1,2}& \textbf{2,4}& \textbf{2,5} & \textbf{1,6} & \textbf{2,7} \\
\hline
\textbf{Key factor}& \textbf{viewpt.}& \textbf{illum.}& \textbf{low light} & \textbf{objects} & \textbf{people} \\
\hline
ORB (stereo)  &  0 &  0 &  0 &  0.742 &  0.995 \\
ORB (RGB-D)   &  0 &  0.764 &  0 &  0.716 &  0.997 \\
DS-SLAM       &  0 &  0 &  0 &  0.994 &  0.996 \\
VINS (color)  &  0 &  0 &  0 &  0.837 &  0 \\
VINS (fisheye)&  0 &  0 &  0 &  0     &  0 \\
InfiniTAMv2   &  0 &  0 &  0 &  0     &  0 \\
\hline
\multicolumn{6}{l}{The values are $\text{C}^{0.3,\infty}\text{S}^{60}\text{-R}$ as defined in Eq. (4)}
\end{tabular}
\label{tab_reloc}
\end{table}

\section{Conclusion}

This paper introduces the OpenLORIS-Scene datasets and metrics for benchmarking lifelong SLAM algorithms. The datasets capture scene changes caused by day-night shifts and human activities, as well as viewpoint changes, moving people, poor illumination, and blur. We found these factors challenging enough to existing SLAM systems. New metrics are proposed to evaluate the localization robustness and accuracy separately. With the datasets and metrics, we hope to help identify shortcomings of SLAM algorithms and to encourage new designs with more robust localization capabilities, such as by introducing high-level scene understanding capabilities. The datasets can also be a testbed of the maturity for real-world deployment of future SLAM algorithms for service robots.

Beyond SLAM, the data may also facilitate a broader scope of long-term scene understanding research for service robots. With proper annotation, it could serve as a benchmark of incremental learning algorithms \cite{she2019openloris}\cite{feng2019challenges}, enabling robots to keep learning new tasks. It would also be interesting to explore spatio-temporal modeling \cite{search_2015_icra} with the data.

\section*{Acknowledgement}

The authors would like to thank Yusen Qin, Dongyan Zhai and Jin Wang for customizing the Segway robot and lending it for this project, and performing odometer-camera calibration. Thank Yijia He for providing LiDAR-camera calibration tools and guidance. Thank Phillip Schmidt, Chuan Chen, Hon Pong Ho and Yu Meng for the technical support of RealSense cameras. Thank Mihai Bujanca and Bruno Bodin for helping integrate OpenLORIS-Scene into SLAMBench. Thank Yinyao Zhang, Long Shi and Zeyuan Dong for helping collect and maintain the data. Thank all the anonymous participants in data collection.




\bibliographystyle{IEEEtran} 
\bibliography{refs} 

\end{document}